\documentclass[10pt,twocolumn,letterpaper]{article}
\usepackage{algorithm}
\usepackage[noend]{algorithmic}

\newcommand{\algname}{FedSelect}

\usepackage[pagenumbers]{cvpr} 

\usepackage[dvipsnames]{xcolor}

\definecolor{cvprblue}{rgb}{0.21,0.49,0.74}
\usepackage[pagebackref,breaklinks,colorlinks,citecolor=cvprblue]{hyperref}
\usepackage{tikz}

\def\paperID{15081} 
\def\confName{CVPR}
\def\confYear{2024}

\title{FedSelect: Personalized Federated Learning with Customized Selection of Parameters for Fine-Tuning}

\author{Rishub Tamirisa$^{\dagger}$, Chulin Xie$^{\dagger,\ddagger}$, Wenxuan Bao$^{\dagger,\ddagger}$, Andy Zhou$^{\dagger}$,  Ron Arel$^{\dagger}$, Aviv Shamsian$^{\S}$ \\
{$^\dagger$Lapis Labs \quad $^\ddagger$University of Illinois Urbana-Champaign \quad $^{\S}$Bar-Ilan University} \\
{\tt\small \{rishubt2,chulinx2,wbao4,andyz3,ronarel2\}@illinois.edu \quad \tt\small aviv.shamsian@live.biu.ac.il
}
}

\begin{document}

\maketitle
\begin{abstract}
Standard federated learning approaches suffer when client data distributions have sufficient heterogeneity. Recent methods addressed the client data heterogeneity issue via personalized federated learning (PFL) - a class of FL algorithms aiming to personalize learned global knowledge to better suit the clients' local data distributions. Existing PFL methods usually decouple global updates in deep neural networks by performing personalization on particular layers (i.e. classifier heads) and global aggregation for the rest of the network. However, preselecting network layers for personalization may result in suboptimal storage of global knowledge. In this work, we propose \textsc{\algname}, a novel PFL algorithm inspired by the iterative subnetwork discovery procedure used for the Lottery Ticket Hypothesis. \textsc{\algname} incrementally expands subnetworks to personalize client parameters, concurrently conducting global aggregations on the remaining parameters. This approach enables the personalization of both client parameters and subnetwork structure during the training process. Finally, we show that \textsc{\algname} outperforms recent state-of-the-art PFL algorithms under challenging client data heterogeneity settings and demonstrates robustness to various real-world distributional shifts.
Our code is available at \href{https://github.com/lapisrocks/fedselect}{https://github.com/lapisrocks/fedselect}.
\end{abstract}

\section{Introduction}

Federated Learning (FL) enables distributed clients or devices to jointly learn a shared global model while keeping the training data local~\cite{mcmahan2017}, which is particularly beneficial for privacy-critical applications~\citep{sheller2020federated}.  Despite its potential, the efficacy of FL is challenged by the inherent heterogeneity of data across different clients.
As the local updates of clients can be remarkably diverse given their heterogeneous local data, the aggregated global model can diverge under the standard FL paradigms~\cite{Li_2020}.

To address this issue, personalized federated learning (PFL) emerges as a promising approach. PFL allows individual clients to maintain their unique, personalized models, with parameters tailored to their local data distributions, while knowledge sharing across different clients. One representative type of PFL algorithm is ``parameter decoupling''. It decomposes the FL model into two distinct components: a global subnetwork shared among all clients (e.g., the feature extractor), and a personalized subnetwork adapted to each client's local distribution (e.g., the prediction head). By personalizing a subset of parameters, parameter decoupling strikes a balance between knowledge sharing among clients and personalization to each client. 

A natural follow-up question would be \textit{how to determine which parameters to personalize}. Previous works typically selected specific layers for personalization, e.g., the prediction head \cite{arivazhagan2019federated,collins2021exploiting}, and the feature extractor \cite{liang2020think,oh2022fedbabu}. However, recent studies \cite{frankle2019lottery,renda2020comparing,frankle2020linear} suggest that even within the same layer, the importance of parameters for prediction can vary significantly. The coarse-grained layer-wise selection of personalized subnetwork may not fully balance knowledge sharing among clients and personalization to each client. Furthermore, current methods typically pre-defining the architecture of the personalized subnetwork, i.e., which layers to personalize, and the architecture of the personalized subnetwork is shared for all clients. These designs limit the performance of parameter decoupling, as the optimal personalized subnetwork often varies depending on each client's specific local data distribution. 

In this paper, we delve into these challenges, exploring a novel strategy named {\textsc\algname} for parameter selection and subnetwork personalization in PFL to unlock the full potential of parameter decoupling. Our method in enlightened by the Lottery Ticket Hypothesis (LTH): deep neural networks contain subnetworks (i.e., winning tickets) that can reach comparable accuracy as the original network. We believe that FL model also contain a subnetwork that is crucial for personalization to a specific client's local distribution, and personalizing that subnetwork can achieve optimal balance between global knowledge sharing and local personalization. Specifically, we believe that parameters that change the most over the course of a local client update should be personalized, while parameters changing the least should be globally aggregated. This is achieved by comparing the element-wise magnitude difference between the state of the client model before and after local training.

Notably, our method is different from the original LTH in twofold: in terms of purpose,  while LTH seeks a sparse network for efficiency, our goal is to enhance personalized FL performance by finding the optimal subnetwork of individual clients for personalization, based on the characteristics of local data distribution.
Methodologically, LTH  prunes less important parameters (i.e., non-winning tickets) to zero value, whereas we assign them the global aggregated value for storing global information, aligning with the collaborative nature of FL. We summarize our contributions below:
\begin{itemize}
    \item We introduce a hypothesis for selecting parameters during training time, aimed at enhancing client personalization performance in FL.
    \item We propose \textsc{\algname}, a novel personalized FL algorithm based on our hypothesis that automatically identifies the customized subnetwork for each client's personalization,  guided by the updating magnitude of each parameter. 
    \item We evaluate  \textsc{\algname} across a range of benchmark datasets: CIFAR10, CIFAR10-C, OfficeHome, and Mini-ImageNet. Our method outperforms state-of-the-art personalized FL baselines in various FL scenarios, encompassing feature and label shifts, as well as different local data sizes. 
    Moreover, our visualizations verify that the learned personalized subnetworks successfully capture the distributional information among clients. 
\end{itemize}

\section{Related Work}
\subsection{Federated Learning} 

There has been extensive work in federated learning on enhancing the training of a global model from various clients with non-independent and identically distributed (non-IID) data \cite{Li2019ASO, aledhari2020federated}. The original approach, FedAvg \cite{mcmahan2017}, aims to develop a single global model computed through the aggregation of local updates from each client without requiring raw client data storage at a central server. However, challenges arise with non-IID data, leading to innovations in optimizing local learning algorithms through objective regularization \cite{acar2021federated}, local bias correction \cite{li2020federated}, and data handling techniques such as class-balanced resampling \cite{hsu2020federated} or loss reweighting \cite{wang2020addressing}. Different from these works, we adapt each client to its local data. 

\subsection{Personalized Federated Learning} 

Personalized federated learning aims to address the issue of data heterogeneity by adapting clients to their local data distribution~\cite{tan2022towards}. Common approaches include multitask learning \cite{Smith2017FederatedML, Agarwal2020FederatedRL}, clustering \cite{mansour2020approaches,Duan2021FlexibleCF, Ghosh2020AnEF}, transfer learning \cite{yu2022salvaging, zhao2018iid}, meta learning \cite{jiang2023improving,fallah2020personalized,chen2019federated}, hypernetworks \cite{shamsian2021personalized}, and gaussian processes \cite{achituve2021personalized}. We focus on partial model personalization, which seeks to improve the performance of client models by altering a subset of their structure or weights to better suit their local tasks. It also addresses the issue of ``catastrophic forgetting'' \citep{McCloskey1989CatastrophicII}, an issue in personalized FL where global information is lost when fine-tuning a client model on its local distribution from a global initialization \citep{Kirkpatrick_2017, pillutla2022federated}. It does this by forcefully preserving a subset of parameters, $u$, to serve as a fixed global representation for all clients. However, existing methods \citep{pillutla2022federated, collins2021exploiting, xu2023personalized} introduced for partial model personalization require hand-selected partitioning of these shared and personalized parameters and choose $u$ as only the input or output layers. 

LotteryFL \citep{li2020lotteryfl} learns a shared global model via FedAvg \citep{mcmahan2017} and personalizes client models by pruning the global model via the vanilla LTH. Importantly, parameters are pruned to zero according to their magnitude after an iteration of batched stochastic gradient updates. However, due to a low final pruning percentage in LotteryFL, the lottery tickets found for each client share many of the same parameters, and lack sufficient personalization \citep{10.1007/978-3-031-19775-8_5}.
\section{Background}
\label{sec:background}
\paragraph{Preliminaries}
\label{sec:preliminaries}
We start by introducing the standard FL setting. Let the set of clients be $C=\{c_1, \dots, c_N\}$, where the total number of clients $N = |C|$. For the $k$-th client $c_k \in C$, the corresponding local dataset is defined as $\mathcal{D}_i = \{x_i^k, y_i^k\}_{i=1}^{N_k}$, where $N_k$ is the number of local data points for client $c_k$. Let $\theta_g$ be the FL global model, and the local loss function (e.g. cross-entropy loss) for client $k$ as $f_k(\theta_g, x)$, we can define the canonical FL objective:
\begin{equation}
\label{eq1}
    \min_{\theta_g}\frac{1}{N}\sum_{k=1}^N\sum_{i=1}^{N_k}f_k(\theta_g, x_i^k)
\end{equation}

\textsc{FedAvg} \citep{mcmahan2017} optimizes the objective in \eqref{eq1} by locally training each of the client models $\theta_k^t$ for $L$  number of local epochs at each communication round $t$. The updated local client model $\theta_k^{t^+}$ is aggregated into the global model $\theta_g^{t+1}$, resulting in the global model update for round $t$ in \textsc{FedAvg} being  $\theta_g^{t+1} \gets \frac{1}{N}\sum_k^N\theta_k^{t^+}$. At the next round, the aggregated global model $\theta_g^{t+1}$ is then redistributed to all clients, resulting in $\theta_k^{t+1} \gets \theta_g^{t+1}, \forall k$.

The goal of personalized federated learning, on the other hand, is to find personalized models $\theta_k$ for each client $k$, either adapted from $\theta_g$ or discovered independently, resulting in a modified objective:

\begin{equation}
\label{eq2}
    \min_{\theta_1, \theta_2, \dots, \theta_N}\frac{1}{N}\sum_{k=1}^N\sum_{i=1}^{N_k}f_k(\theta_k, x_i^k)
\end{equation}

Partial model personalization refers to the procedure in which model parameters are partitioned into shared and personalized parameters, denoted $u$ and $v$, for averaging and personalization. We consequently define $\theta_k = (u_k, v_k)$, where $u$ denotes a set of shared global parameters (e.g., aggregated via \textsc{FedAvg}), and $v_k$ the personalized client parameters. Substituting the tuple $(u_k, v_k)$ for each client model $\theta_k$ yields the partial model personalization objective as in \citep{pillutla2022federated}.

\section{\algname}

In this section we discuss the motivation for our algorithm, followed by a description of the top-level procedure (Algorithm \ref{alg:{\algname}}). We then provide detailed overviews of 2 sub-procedures within \textsc{\algname} (Algorithms \ref{alg:LocalAlt}, \ref{alg:GradSelect}). We present an analogy between \textsc{\algname} and the algorithm used by \citep{li2020lotteryfl} for finding winning lottery ticket networks, and note important distinctions in our application to FL.

\subsection{Motivation}
\label{sec:motivation}

Prior works in FL selectively personalize specific layers of the  model to better learn the heterogeneous distributions of local clients~\cite{lee2023surgical, pillutla2022federated, liang2020think,lifedbn, collins2021exploiting}, which are  referred to as performing ``parameter-decoupling''.
For example, they rely on the conventional wisdom that the final linear layer of deep neural networks (DNNs) for classification contains a more enriched semantic representation with respect to the output prediction neurons \citep{Yosinski2014HowTA,arivazhagan2019federated}, which could be more suitable for personalization. 

In this work, we propose a novel hypothesis describing that only \textit{parameters} that change the most during training are necessary as personalized parameters for client models during FL; the rest can be aggregated as global parameters. Analogous to the sparse winning ticket networks found via the LTH, \textsc{\algname} aims to elicit an optimal subnetwork for \textit{fine-grained personalization} of individual client model. Our primary intuition is that drastic distributional changes in the client personalization task may be better accommodated by preserving the accumulated global knowledge and personalized knowledge in a \textit{parameter-wise} granularity, rather than \textit{layer-wise}. Following this we state our hypothesis: 

\paragraph{FL Gradient-based Lottery Ticket Hypothesis.} \textit{When training a client model on its local distribution during federated learning, parameters exhibiting minimal variation are suitable for encoding shared knowledge, while parameters demonstrating significant change are optimal for fine-tuning on the local distribution and encoding personalized knowledge.}

Now we introduce some basic notations regarding neural subnetwork and vanilla LTH.
Given a neural network with parameters $\theta$, we define a subnetwork via a binary mask $m$. For convenience, we use the Hadamard operator $\odot$ to be an indexing operator for $m$, rather than an elementwise multiply operator. For example, $\theta \odot m$ assumes $\theta$ and $m$ have the same dimensions and returns a reference to the set of parameters in $\theta$ where $m$ is not equal to zero. We also define the operator $\neg$ to invert the binary masks.
To identify winning tickets via the vanilla LTH, the iterative magnitude search (IMS) procedure is proposed \citep{frankle2019lottery}, which finds a mask $m$ for network parameters $\theta$ such that $\theta \odot m$ can be trained in isolation to match the performance of $\theta$ on a given dataset. For clarity, we restate the IMS procedure as follows. Given a pruning rate $p$, an initial model $\theta$ is trained for $j$ iterations, yielding $\theta^+$. Next, the smallest $p\%$ of parameters in $\theta^+$ are identified via a binary mask $m$. The original model $\theta$ is then pruned, given by $\theta \odot \neg m \gets 0$, and the process repeats until the desired sparsity of $m$ is achieved.

 While we draw inspiration from the winning ticket IMS, there are three distinct and notable differences in our approach that set it apart, 
 which will be covered in greater detail in Sections~\ref{sec:client_update} and \ref{sec:subnet_rep}.  First, rather than iteratively finding smaller subnetworks, our approach iteratively \textit{grows} a subnetwork within each client. Second, we do not perform any parameter pruning; instead, we use $m_k$ for each client $\theta_k$ to obtain the partition of global and personalized parameters as $(u_k, v_k) = (\theta_k \odot \neg m_k, \theta_k \odot m_k)$. Third, the mask pruning update to $m_k$ is computed based on the element-wise magnitude of parameter update $\nabla v_k$ from local client training rather than the magnitude of parameter $v_k$ itself. Regarding the winning ticket found via our IMS, the set of parameters selected in \textsc{\algname} will permanently remain personalized (i.e., always kept local and not sent to server); this set will grow in size over the course of FL.

\begin{figure*}
  \centering
    \vspace{-0.2in}
    \includegraphics[width=0.90\linewidth]{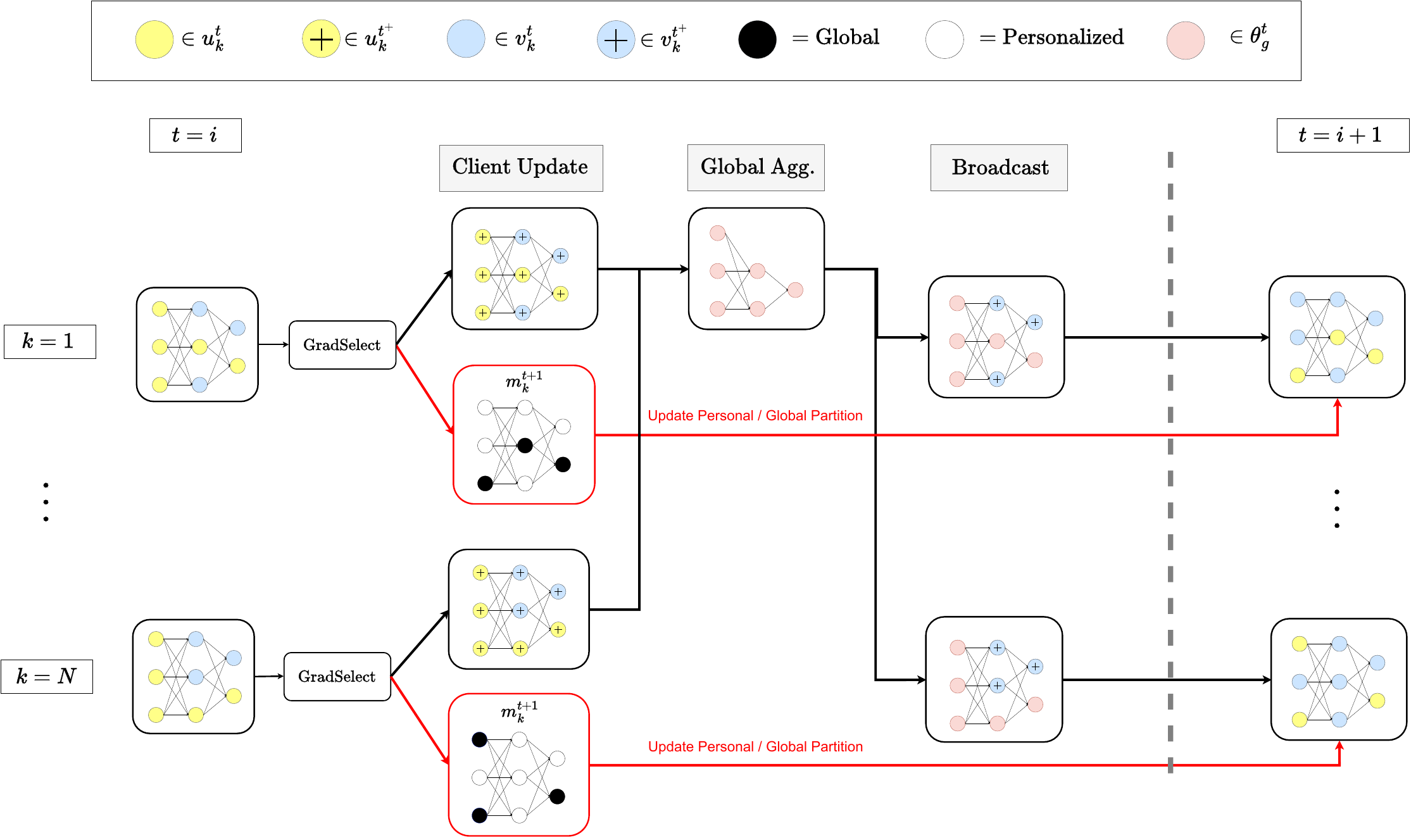}
  \caption{Illustration of the \textsc{\algname} algorithm. An example subnetwork update for communication round $t=i$ into $t=i+1$ is depicted for $N$ clients, where 2 clients are shown. There are 4 key steps: (1) the local update / new partition via GradSelect (Algorithm \ref{alg:GradSelect}), (2) the aggregation of global parameters $v_k^t$, (3) broadcast of global parameters to the updated clients, and (4) application of the new mask $m_k^{t+1}$ as a partition for the global/personalized parameters of each client in the subsequent round $t+1$. In our algorithm, $u_k^t$ denotes the global parameters for each client at round $t$; $v_k$ denotes the personalized parameters for each client at round $t$; $u_k^{t^+}$ denotes the updated global parameters after GradSelect; $v_k^{t^+}$ denotes the updated personalized parameters after GradSelect;  $m_k^{t+1}$ is the binary mask with ``0'' denoting global parameter and ``1'' denoting personalized parameter; $\theta_g^t$ is the aggregate global parameters materialized each round.}
  \label{fig:illustration}
  \vskip -5pt
\end{figure*}

There are two additional considerations when designing an algorithm that selects parameters using our hypothesis for personalization while training during FL: (1) computational efficiency for subnetwork discovery and (2) a suitable mechanism for performing a local update on both $u_k$ and $v_k$. A possible algorithm for selecting $v_k$ for each communication round could involve performing federated averaging, followed by a gradient-based subnetwork search using a modified version of the IMS that computes $m$ using $\nabla v_k$. However, this would require performing $j \times L$ iterations of local training, which would incur an undesirable computational overhead during FL. Next, we discuss how \textsc{\algname} addresses both points (1) and (2).


 \textsc{\algname} (Algorithm \ref{alg:{\algname}}) takes the following as input: the set of clients $C$, an initial model $\theta_I$, the number of communication rounds  $T$, the personalization rate $p$, and the personalization limit $\alpha$. We reuse the notation from Section~\ref{sec:background} to refer to global and personalized parameters as either $u$ and $v$, or positions in the client masks $m_k$ equal to $0$ or $1$, respectively. 

 \subsection{Client Update Overview} 
 \label{sec:client_update}
 At the beginning of FL, all client models are initialized with the same global model $\theta_g^0$, denoted via binary client masks set as $m_k^0 = 0$ applied to each client model $\theta_k^0$. Here, we use GradSelect as a black-box procedure for obtaining a personalized subnetwork update, which will be described in further detail in Section~\ref{sec:gardselect}. First, given the current communication round $t$, GradSelect updates each clients' current global parameters ($u_k^t$) and personalized parameters ($v_k^t$); these are identified using the current mask $m_k^t$ as an index into $\theta_k^t$. Then, the next set of parameters to include for personalization in $v_k^t$ are chosen according to a scalar personalization rate $p \in [0, 1]$. This is notated via values at the corresponding indices for $v_k^t$ in $m_k^t$ being set to $1$. Importantly, in each communication round, the new parameters in $v_k^t$ are selected solely from the set of global parameters that have been aggregated such that $\text{Idx}(v_k^0) \in \text{Idx}(v_k^1) \in ... \in \text{Idx}(v_k^T)$. We note a subtlety in our formulation in Eq. \ref{eq2} that differs from the original PFL objective proposed in \citep{pillutla2022federated}. Namely, in the original partial model personalization framework, it is expected that all client models share the same global parameters denoted as $u^t$. In \textsc{\algname}, clients may learn different personalized parameters $v_k^t$, resulting in \textit{differing partitions of the shared global parameters} $u_k^t$.

\subsection{Global Aggregation} 
\label{sec:golbal_agg}
After each client performs its update, the global parameters $u_k$ for each client are aggregated. Since $m_k$ is likely to be heterogeneous across clients in typical personalized FL settings, averaging the global parameters $u_k$ for each client requires careful handling. This is because the locations of global parameters (values of $0$ in $m_k$) may be non-overlapping for multiple parameter indices. In \textsc{\algname}, global averaging for a given index in the current global model $\theta_g^t$ only occurs across parameters $u_k$ in each client for which the corresponding mask entry in $m_k = 0$. We use $\omega_t$ as a tool of implementation to store the number of clients contributing to each global parameter in $\theta_g^t$ element-wise. By this construction, different subsets of clients $c_k \in C$ can contribute to different global parameters in $\theta_g^t$. We provide a visual description of this procedure in Figure \ref{fig:illustration}. As a result, the first communication round in \textsc{\algname} where all clients masks are set to $0$ performs pure federated averaging. A consequence of this formulation is that in the extreme case where global parameters $u_k$ is completely disjoint across all clients, each client will effectively undergo purely local training. However, full disjointedness across $u_k$ is unlikely in typical FL settings where clients are likely to have similar data distributions. We hypothesize that clients with similar data distribution trained using \textsc{\algname} will learn similar personalized subnetworks; we show visualizations of these correlations in Section~\ref{sec:results}. 

\begin{figure}[tb]
\vskip -10pt
\setlength{\textfloatsep}{5mm}
\begin{algorithm}[H]
   \caption{{\algname}}
   \label{alg:{\algname}}
\begin{algorithmic}
   \STATE {\bfseries Input:} $C = \{c_1, \dots, c_N\}, \theta_I, T, L$
   \STATE {\bfseries Server Executes:}
   \begin{ALC@g}
        \STATE Initialize all client models $\{\theta_i^0\}_{i=1}^N$ with $\theta_I$
        \STATE Initialize all client masks $\{m_i^0\}_{i=1}^N$ with $0$s
        \FOR {each round $t$ in $0,1,\dots,R-1$}
        
            \FOR {each client $c_k \in C$ {\bfseries in parallel}}
            \STATE \textit{\textcolor{teal}{\# Executed locally on client $c_k$}}
                \STATE $u_k^t \gets \theta_k^t \odot \neg m_k^t$
                \STATE $v_k^t \gets \theta_k^t \odot m_k^t $
                \STATE $u_k^{t^+}, v_k^{t^+}, m_k^{t+1} \gets$ GradSelect$(u_k^t, v_k^t)$
                
            \ENDFOR
        \STATE \textit{\textcolor{teal}{\# Averaging occurs only across clients where the mask is $0$ for a given parameter's position}}
        \STATE $\theta_g^t \gets \textbf{0}$
        \STATE $\omega^t \gets \textbf{0}$
        \FOR {$k=1$ {\bfseries to} $N$}
            \STATE $\theta_g^t \odot \neg m_k \gets (\theta_g^t \odot \neg m_k^t) + u_k^{t^+}$
            \STATE $\omega^t  \odot \neg m_k \gets (\omega^t \odot \neg m_k^t) + \neg m_k^t$
        \ENDFOR
        \STATE $m_g^t \gets$ Binary mask for $\omega^t \neq 0$
        \STATE $\theta_g^t \odot m_g^t \gets \frac{\theta_g^t \odot m_g^t}{\omega^t  \odot m_g^t}$
        \FOR {$k=1$ {\bfseries to} $N$}
            \STATE \textit{\textcolor{teal}{\# Distribute global params to clients' non-selected params, located via $\neg m_k^t$}}
            \STATE $\theta_k^{t+1} \odot \neg m_k^{t} \gets \theta_g^t \odot \neg m_k^{t}$
            \STATE $\theta_k^{t+1} \odot m_k^{t} \gets {v_k^{t^+}}$
        \ENDFOR
        \ENDFOR
    \end{ALC@g}
\end{algorithmic}
\end{algorithm} 

\vskip -10pt
\end{figure}

\begin{figure}[tb]
\vskip -10pt
\setlength{\textfloatsep}{5mm}
\begin{algorithm}[H]
   \caption{{LocalAlt}$(u_{k, 0}, v_{k, 0})$ \cite{pillutla2022federated}}
   \label{alg:LocalAlt}
\begin{algorithmic}
    \STATE {\bfseries Input:} {\text{Global/personalized parameters} $u_{k, 0}, v_{k, 0}$, \text{\# of steps} $\tau$, \text{global/local learning rates} $\gamma_v, \gamma_u$, \text{Batched data} $\mathcal{D} = \{b_0, b_2, \dots, b_{\tau-1}\}$}
        
    \FOR {$i = 0, 1, \dots, \tau - 1$}
        \STATE $v_{k, i+1} \gets v_{k, i} - \gamma_v \nabla_{v_k}f_k((u_k, v_{k, i}), b_i)$
    \ENDFOR
    \STATE $v_{k}^+ \gets v_{k, \tau}$
    \FOR {$i = 0, 1, \dots, \tau - 1$}
        \STATE $u_{k, i+1} \gets u_{k, i} - \gamma_u \nabla_{u_k}f_k((u_k, v_{k}^+), b_i)$
    \ENDFOR
    \STATE $u_{k}^+ \gets u_{k, \tau}$
    \STATE {\bfseries Return} $v_{k}^+, u_{k}^+$
\end{algorithmic}
\end{algorithm}

\vskip -10pt

\setlength{\textfloatsep}{5mm}
\begin{algorithm}[H]
   \caption{{GradSelect}$(u_k, v_k)$}
   \label{alg:GradSelect}
\begin{algorithmic}
    \STATE {\bfseries Input:} {\text{Global/personalized parameters} $u_k$, $v_k$, \text{Per. rate} $p$, \text{\# local epochs} $L$, \text{Per. bound} $\alpha$ }
    \STATE $u_{k, 0} \gets u_k$
    \STATE $v_{k, 0} \gets v_k$
    \FOR {$i = 0, 1, \dots, L - 1$}
        \STATE $u_{k, i + 1}, v_{k, i + 1} \gets$ LocalAlt($u_{k, i}, v_{k, i}$)
    \ENDFOR
    \IF {sparsity of $\neg m_k < \alpha$}
        \STATE $\Delta u_k \gets |u_{k, L} - u_{k, 0}|$
        \STATE $m_k^+ \gets$ binary mask for largest $p\%$ values in $\Delta u_k$
        \STATE \textit{\textcolor{teal}{\# Element-wise Binary OR of $m_k^+$ and $m_k$}}
        \STATE $m_k^+ \gets m_k^+ \vee m_k$ 
    \ELSE
        \STATE $m_k^+ \gets m_k$
    \ENDIF
    \STATE {\bfseries Return} $u_{k, L}, v_{k, L},  m_k^+$
\end{algorithmic}
\end{algorithm}

\vskip -10pt
\end{figure}

\subsection{Subnetwork Representation} 
\label{sec:subnet_rep}
A key property of \textsc{\algname}, mentioned in Section~\ref{sec:motivation}, is that the personalized subnetwork representation gradually grows in size as the communication rounds progress. Specifically, the subnetwork size grows by a factor of $p\%$ until the corresponding client mask $m_k$ reaches the maximum subnetwork mask sparsity $\alpha$, a scalar defined within $[0, 1]$. We also refer to $\alpha$ as the personalization limit, the central hyperparameter of our proposed algorithm, where larger $\alpha$ indicates greater personalization. Therefore, when $\alpha=1.0$, \textsc{\algname} computes local-only training of each client after $m_k$ for each client reaches sparsity $\alpha$. Conversely, when $\alpha = 0$, \textsc{\algname} reduces exactly to \textsc{FedAvg}. We consider the behavior of our proposed algorithm as enabling a ``rough interpolation'' between these two extremes for personalization.

Furthermore, for $\alpha > 0.50$ and sufficiently many communication rounds for the chosen personalization rate $p$, a majority of parameters in all client models will be selected for personalization, resulting in decreased communication costs over time within \algname.

By our design of evolving masks over the course of FL rather than within a single communication round, \textsc{\algname} achieves a similar time complexity to other alternating minimization-based personalized FL methods like FedRep \citep{collins2021exploiting} and FedPAC \citep{xu2023personalized}, while performing a first-of-its-kind selection of personalized parameters.

\subsection{GradSelect}
\label{sec:gardselect}

We now describe GradSelect in isolation. GradSelect takes as input a partition of the current global parameters $u_k^t$ and personalized parameters $v_k^t$ at communication round $t$, a personalization rate $p$, and personalization limit $\alpha$. The goal is to compute an update to the client model that grows the current subnetwork by a factor of $p\%$ while also training the current client model. To this end, we apply our FL Gradient-based Lottery Ticket Hypothesis to both train the client model's parameters and discover new parameters for personalization. First, to update both $u_k$ and $v_k$ for each client, we use LocalAlt (Algorithm \ref{alg:LocalAlt}) from the partial model personalization framework \citep{pillutla2022federated}. LocalAlt was introduced to update a predefined set of shared and personalized parameters, $u_k$ and $v_k$, by alternating full passes of stochastic gradient descent between both sets of parameters.
        
The LocalAlt training epochs within GradSelect are analogous to the training iterations used to compute lottery ticket updates in the aforementioned IMS procedure. Continuing the analogy, we store the state of the client parameters before LocalAlt $\theta_k$, and compute the absolute value of the change of all global parameters after LocalAlt in $\Delta u_k$. Taking the largest $p\%$ of values in $\Delta u_k$, we create a the new mask $m_k^+$, which finishes the subnetwork update. Because $m_k^+$ is the mask to be used for local updates in the subsequent communication round, GradSelect also returns the newly-trained global/personalized parameters from the current mask partition $m_k$ given by $u_{k,L}, v_{k,L}$. The parameter $p$ controls the rate at which personalized subnetworks grow, and $\alpha$ controls the maximum size of the subnetwork. We include the result of varying $\alpha$ in Section~\ref{sec:results}, as well as the effect of changing $p$ in Appendix B.

\vspace{-1mm}
\section{Experiments}
\vspace{-1mm}
In this section, we compare \algname{} with different approaches from FL and personalized FL. We use
a variety of datasets and learning setups to demonstrate the superiority of \algname{}. We will make our source code publicly available to encourage reproducibility. Additional experimental results and details are provided in Appendix B.
\subsection{Experimental Setup}

\paragraph{Models \& Datasets.} We show results for our experimental setting across a breadth of benchmark datasets: CIFAR-10 \citep{Krizhevsky2009LearningML}, CIFAR10-C \citep{hendrycks2019robustness}, Mini-ImageNet \citep{vinyals2017matching}, and the OfficeHome dataset \citep{Venkateswara2017DeepHN}. These settings additionally cover label shifts and feature shifts, which are common distributional shifts in real-world data. We use a ResNet18 \citep{he2015deep} backbone with random initialization on CIFAR-10, CIFAR10-C, and Mini-ImageNet. For OfficeHome, we follow \cite{sun2021partialfed} to use a ResNet18 backbone pretrained on ImageNet~\cite{5206848}.

\vspace{-1mm}
\paragraph{Baselines.}  We compare our method to the \textit{full model personalization} methods, including local-only training,  FedAvg~\cite{mcmahan2017} with local fine-tuning (FedAvg + FT),  Ditto \cite{li2021ditto}, as well as \textit{partial model personalization} methods, including  FedPAC \cite{xu2023personalized}, FedBABU \cite{oh2022fedbabu}, FedRep \cite{collins2021exploiting}, FedPer \cite{arivazhagan2019federated}, and LG-FedAvg \citep{liang2020think}. 
\vspace{-1mm}
\paragraph{FL Settings \& Hyperparameters.} We consider the typical cross-silo setting \citep{liu2022privacy}  with dozens of clients and full-client participation. For the CIFAR-10 and CIFAR-10C experiments, the number of clients $N = |C| = 10$, with each client assigned $s=2$ classes. In the Mini-ImageNet experiment, we set $N = 20$, with $s=10$ classes assigned to each client. Finally, the OfficeHome experiment involves $N = 4$ clients, one for each of the $4$ domain shifts in the dataset; each client is allocated all 65 classes. We vary the personalization limit, $\alpha$, within $[0.05, 0.30, 0.50, 0.80]$. Further details on hyperparameter tuning for both \textsc{\algname{}} and the compared baselines are provided in Appendix A.


\vspace{-1mm}
\paragraph{Evaluation Metric.} For all methods, the mean accuracy of the final model(s) across individual client data distributions calculated at the final communication round is reported. For FedAvg, accuracy is reported for a single global model. However, for other methods that learn personalized client models, the final average accuracy is reported by averaging individual clients' personalized model accuracies. 
\vspace{-1mm}
\paragraph{Training Settings.} 
We use standard SGD for performing local client optimization for all methods based on their respective training objective. 
To fairly compare the performance of these methods, we fix the number of local training epochs across all methods to $3$. The number of communication rounds $T$ is set to $200$ for all experiments except OfficeHome, where $T=30$. For both CIFAR-10 and CIFAR10-C, each client is given 100 training samples. For Mini-ImageNet, we use 20\% of the total training data, sampled for 20 clients in a non-iid manner.
The 4 clients in the OfficeHome experiment were given the full training partition for each of their corresponding domains, resulting in about $2,000$ training samples per client. Further details on our data partition are provided in Appendix A.

\subsection{Results} 
\label{sec:results}

\begin{figure*}
  \vspace{-0.2in}
  \centering
    \includegraphics[width=1.0\linewidth]{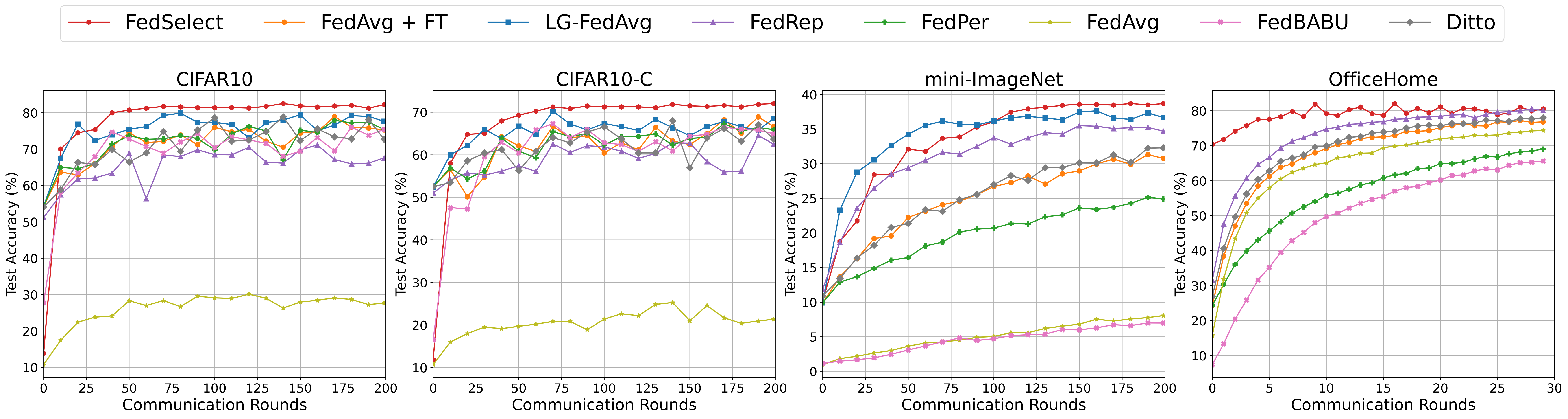}
  \caption{Test accuracy across communication rounds of \textsc{\algname} and baselines under the experimental settings in Table \ref{table:table1}. \textsc{\algname} outperforms all baselines and exhibits more stable convergence.}
  \label{fig:convergence}
     \vspace{-3mm}  
\end{figure*}

\begin{table}
  \scriptsize
  \centering

    \begin{tabular}{@{}lcccc@{}}
    \toprule
    Method & CIFAR10 & CIFAR10-C & Mini-ImageNet & Officehome \\
    \midrule
    Local Only & 74.60 & 69.50 & 34.98 & 75.60 \\
    FedAvg & 27.70 & 21.35 & 8.06 & 74.35 \\
    FedAvg + FT & 75.30 & 66.65 & 30.78 & 76.84 \\
 Ditto & 72.75 & 63.70 & 32.30 & 77.99 \\
     \midrule
    FedPAC & 77.20 & 69.50 & 37.72 & 66.61 \\
    FedRep & 67.60 & 62.50 & 34.71 & 80.07 \\
    FedPer & 75.40 & 65.95 & 24.90 & 69.01 \\
    FedBABU & 75.45 & 64.95 & 6.99 & 65.63 \\
    LG-FedAvg & 77.65 & 68.55 & 36.68 & 78.75 \\
    \midrule
    \algname{} & \textbf{82.25} & \textbf{72.05} & \textbf{38.69} & \textbf{80.51} \\
 
    \bottomrule
  \end{tabular}
  \caption{Personalized accuracy of different methods on four datasets. \textsc{\algname{}} achieves the highest personalized performance.}
\vspace{-3mm}  
  \label{table:table1}
\end{table}

\begin{table}
  \centering
  \scriptsize
  \setlength{\tabcolsep}{1.0mm}{
  \begin{tabular}{@{}lcccc@{}}
    \toprule
    Method & CIFAR10 & CIFAR10-C & Mini-ImageNet & Officehome \\
    \midrule
    \algname{} ($\alpha=0.05$) & 80.10 & 68.30 & 37.60 & 78.46 \\
    \algname{} ($\alpha=0.30$) & \textbf{82.25} & 70.60 & 37.84 & \textbf{80.51} \\ 
    \algname{} ($\alpha=0.50$) & 82.20 & \textbf{72.05} & \textbf{38.69} & 76.08 \\
    \algname{} ($\alpha=0.80$) & 81.40 & 71.40 & 34.51 & 76.65 \\
    \bottomrule
  \end{tabular}
  }
  \caption{Performance of \textsc{\algname{}} when varying the personalization limit $\alpha$.}
  \label{tab:table2}
   \vspace{-4mm}
\end{table}

\paragraph {Personalization Performance. } We report the results on four datasets in Table \ref{table:table1}. In all cases, \textsc{\algname{}} achieves state-of-the-art results, surpassing baseline methods with an average improvement of over 2.4\%. It is also clear that our method is scalable to various feature/label shifts, evidenced by consistent performance across CIFAR10-C and Mini-ImageNet experiments. In contrast, the performance of methods such as FedBABU, FedPer, FedRep, and Ditto degrades significantly for these label/feature-shift experiments. We also note that FedAvg with fine-tuning (FedAvg + FT) performs competitively with other PFL baselines, which has also been previously observed in \citep{xu2023personalized, collins2021exploiting}. 
The consistent superiority of \textsc{\algname{}} demonstrates the benefits of learning \textit{which} parameters to personalize, \textit{while} fine-tuning them. We also observe in the test accuracy convergence plots in Figure \ref{fig:convergence} that our method converges more smoothly to its final test accuracy than the other baselines. Early convergence in FL is useful for enabling early stopping and preventing further communication costs.

\begin{table}
  \centering
  \scriptsize
    \setlength{\tabcolsep}{1.0mm}{
    \begin{tabular}{@{}lcccc@{}}
    \toprule
    Variant & CIFAR-10 & CIFAR10-C & Mini-ImageNet & OfficeHome\\
    \midrule 
    Personalize Least & 79.87 & 64.20 & 35.65 & 78.23\\
    Layer A & 79.96 & 65.76 & 35.43 & 77.06 \\
    Layer B & 78.36 & 62.81 & 32.74 & 79.08 \\
    Layer C  & 79.24 & 64.82 & 33.33 & 79.64  \\
    Layer D & 78.08 & 62.50 & 34.91 & 79.11 \\
    Random & 79.73 & 61.12 & 33.28 & 76.33\\
    \algname{} ($\alpha=0.30$) & \textbf{82.25} & 70.60 & 37.84 & \textbf{80.51}   \\
    \algname{} ($\alpha=0.50$) & 82.20 & \textbf{72.05} & \textbf{38.69} & 76.08  \\
    \bottomrule
  \end{tabular}
  }
  \caption{Ablation study for three variants: Personalize Least refers to the inverse of our hypothesis (personalize parameters with the least variation); Layer A/B/C/D refers to personalizing specific internal layers of ResNet18; Random refers to choosing a random partition between global and personal parameters.}
  \label{table:table3}
     \vspace{-3mm}  
\end{table}

\begin{figure}[t]
  \centering
   \includegraphics[width=0.9\linewidth]{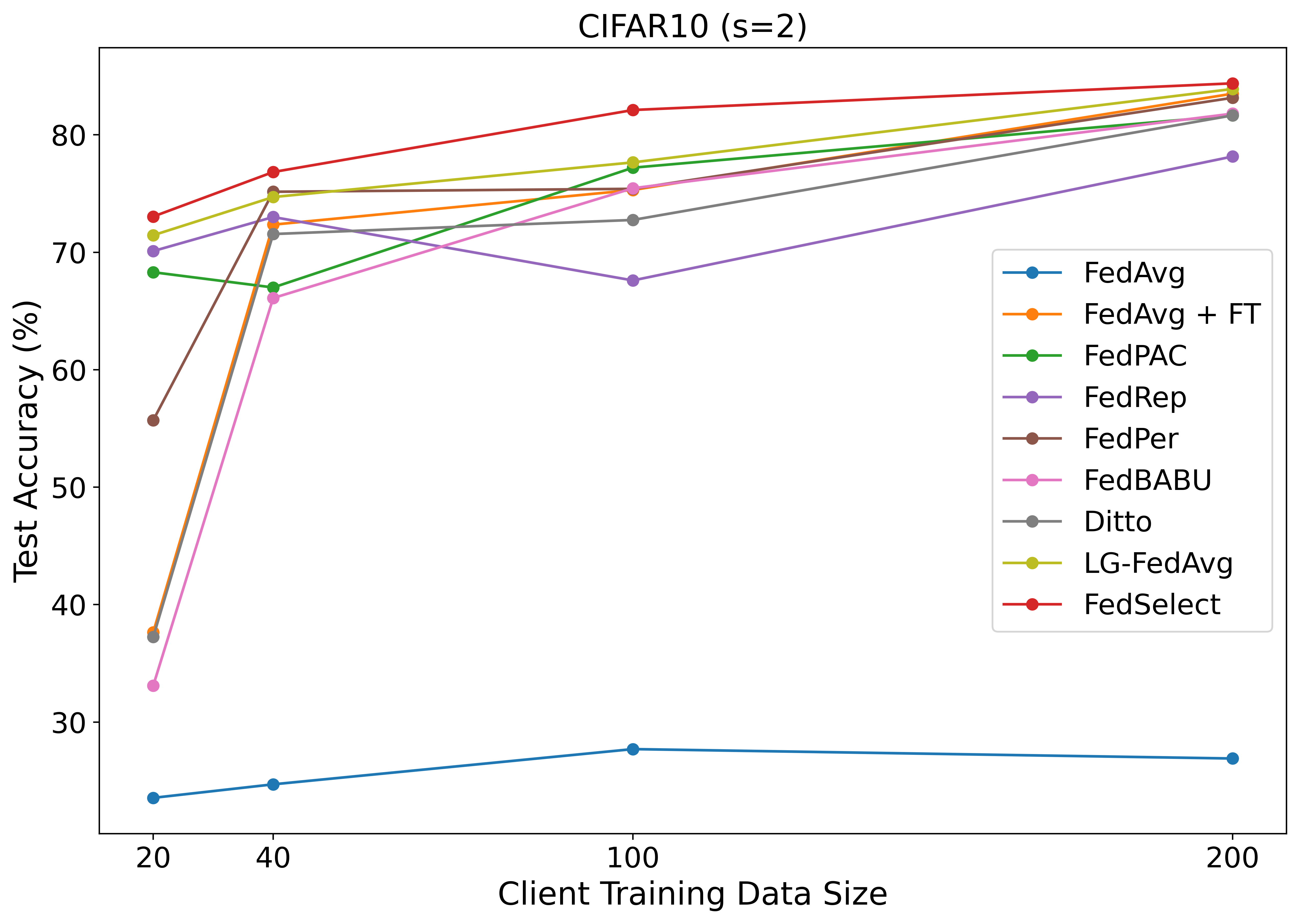}
   \vspace{-3mm}  
   \caption{Personalized performance on CIFAR-10 with different local training data size and shard $s=2$. \textsc{\algname{}} outperforms prior methods.}
   \label{fig:nc}
      \vspace{-3mm}  
\end{figure}

\paragraph {Effect of Personalization Limit $\alpha$.} In Table \ref{tab:table2}, we find that adjusting the personalization limit $\alpha$ enables the performance 
of \textsc{\algname{}} to be tuned under different client data distributions. Recall from Section~\ref{sec:subnet_rep} that the two extremes of \textsc{\algname{}}, $\alpha=0$ and $\alpha=1.0$, perform FedAvg and eventual local training, respectively. The results in Table \ref{tab:table2} suggest that choosing a middle-ground between personalizing and globally averaging parameters is beneficial. In particular, we recommend $\alpha \in [0.3, 0.5]$ as suitable for most heterogenous client distributions. We present results for other values of $\alpha$ in Appendix B.

\begin{figure*} 
  \vspace{-0.2in}
  \centering
    \includegraphics[width=1.0\linewidth]{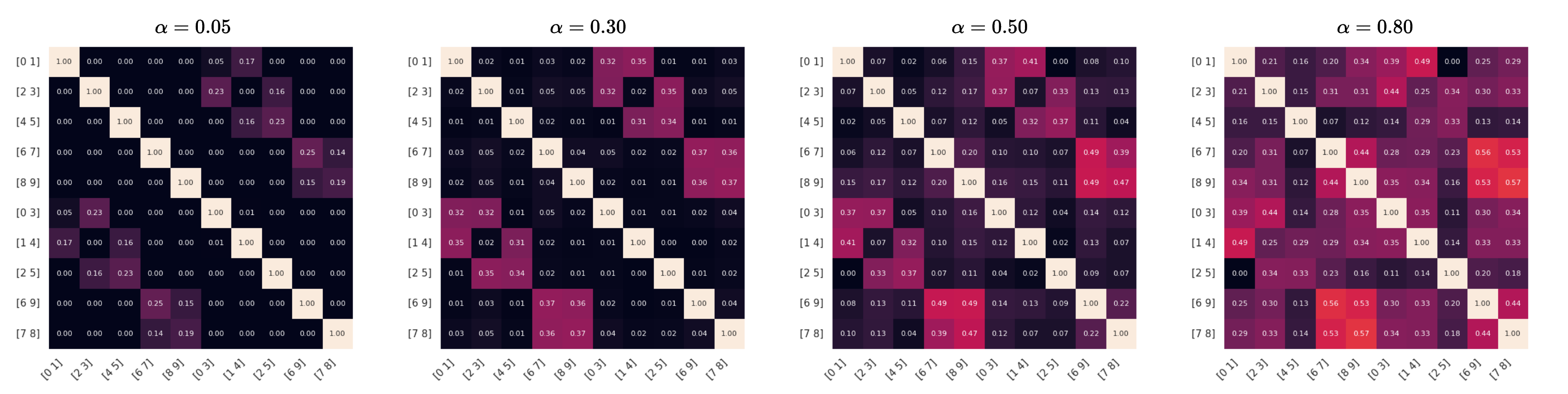}
    \vspace{-3mm}
  \caption{Normalized intersection-over-union (IoU) overlap of the subnetwork masks $m_k$ in the final round in the ResNet18 final linear layer for each client in the CIFAR10 experiments from Table \ref{tab:table2}. Increasing $\alpha$ is shown from left-to-right. Each client was assigned 2 classes; the class labels are shown along the rows and columns of each matrix. Clients with similar labels develop similar subnetworks; increasing $\alpha$ results in more personalized parameters, but less distinct subnetworks.}
  \label{fig:iou}
\end{figure*}

\paragraph {Effect of Training Data Size.} We conduct an additional set of experiments on CIFAR-10 to demonstrate the performance of \textsc{\algname{}} and the aforementioned baseline methods as the size of training data increases. The purpose of this experiment is to stress-test the performance of the algorithms under significantly limited data settings, as well as consistency in performance as the client training sets scale in size. The results shown in Figure \ref{fig:nc} showcase the robustness of \textsc{\algname{}} to a setting with significantly fewer data samples, whereas other baselines like FedBABU and FedPer degrade to as little as 33.1\% and 55.7\%, respectively. Full table results for this experiment are included in Appendix B.

\paragraph {Ablation Study.} We perform ablation experiments to verify the design choices behind our proposed algorithm. Depicted in Table \ref{table:table3}, we test the following three variants of parameter-wise FL algorithms. First, we address the inverse of our hypothesis, which is to personalize parameters that change the least (Personalize Least). Second, we test 4 different layers (denoted A, B, C, \& D) in ResNet18 for individual personalization. In the case of the layer-wise study, we also note that baseline algorithms like FedRep and FedBABU are versions of parameter-wise PFL where the final linear layer is decoupled. We provide further details on the set of layers studied in Appendix B. Third, we test a random partition of personal vs. global parameters. In Table \ref{table:table3}, we observe that all ablations perform worse than \textsc{\algname{}}, which further supports our FL Gradient-based Lottery Ticket Hypothesis.

\paragraph {Subnetwork Development During FL.} While raw test accuracy of the final communication rounds provides useful insights into the performance of \textsc{\algname{}}, we also seek to visualize the behavior of the clients' collaboration of both parameters and subnetworks. In Figure \ref{fig:iou}, we showcase the similarity of subnetwork structures across clients based on their label distributions. We observe that clients that share at least one label have exhibit significant overlap in their subnetwork masks. We also note that the increased overlap in subnetwork parameters due to increasing $\alpha$ results in less distinct but more locally trained parameters.

\section{Limitations}
\vspace{-1mm}
In this work, we mainly focus on improving the personalization performance of FL.
Nevertheless, personalized federated learning faces challenges in balancing personalization and model generalization (e.g., test-time distribution shifts), as local updates may lead to overfitting and biased outcomes. Heterogeneous datasets contribute to diverse knowledge for FL, while security risks such as adversarial attacks and model poisoning persist. The decentralized nature introduces communication overhead and resource demands, impacting scalability and real-time responsiveness. 
Ongoing research is crucial to address these limitations and strike a balance between personalization, efficiency,  model robustness, and privacy in federated learning systems.

\vspace{-1mm}
\section{Conclusion}
\vspace{-1mm}
In this work, we propose \algname{}, an approach that adaptively selects model parameters for personalization while concurrently conducting global aggregations on the remaining parameters in personalized federated learning. \algname{} represents a significant stride towards achieving a harmonious balance between individual customization and collective model performance while reducing communication costs. By dynamically tailoring specific parameters to local data characteristics, \algname{} mitigates the risk of overfitting and enhances personalization. Simultaneously, its global aggregation mechanism ensures the model maintains robust and generalized performance across the entire federated network. Finally, we evaluated \algname{} on multiple datasets with different learning setups and showed that it outperforms previous approaches by a significant margin. The impressive performance of \algname{} paves the way for intriguing future research directions in this domain.
\\

\textbf{Acknowledgements.} 
We thank Bo Li for the initial discussion and constructive suggestions. 
We also thank the National Center for Supercomputing Applications (NCSA) and Illinois Campus Cluster Program (ICCP) for supporting our computing needs. This work used NVIDIA GPUs at NCSA Delta
through allocations CIS230117 from the Advanced Cyberinfrastructure Coordination
Ecosystem: Services \& Support (ACCESS) program, which is supported by NSF Grants \#2138259,
\#2138286, \#2138307, \#2137603, and \#2138296.

\newpage
{
    \small
    \bibliographystyle{ieeenat_fullname}
    \bibliography{main}

\begin{thebibliography}{47}
\providecommand{\natexlab}[1]{#1}
\providecommand{\url}[1]{\texttt{#1}}
\expandafter\ifx\csname urlstyle\endcsname\relax
  \providecommand{\doi}[1]{doi: #1}\else
  \providecommand{\doi}{doi: \begingroup \urlstyle{rm}\Url}\fi

\bibitem[Acar et~al.(2021)Acar, Zhao, Matas, Mattina, Whatmough, and Saligrama]{acar2021federated}
Durmus Alp~Emre Acar, Yue Zhao, Ramon Matas, Matthew Mattina, Paul Whatmough, and Venkatesh Saligrama.
\newblock Federated learning based on dynamic regularization.
\newblock In \emph{International Conference on Learning Representations}, 2021.

\bibitem[Achituve et~al.(2021)Achituve, Shamsian, Navon, Chechik, and Fetaya]{achituve2021personalized}
Idan Achituve, Aviv Shamsian, Aviv Navon, Gal Chechik, and Ethan Fetaya.
\newblock Personalized federated learning with gaussian processes.
\newblock \emph{Advances in Neural Information Processing Systems}, 34:\penalty0 8392--8406, 2021.

\bibitem[Agarwal et~al.(2020)Agarwal, Langford, and Wei]{Agarwal2020FederatedRL}
Alekh Agarwal, John Langford, and Chen-Yu Wei.
\newblock Federated residual learning.
\newblock \emph{ArXiv}, abs/2003.12880, 2020.

\bibitem[Aledhari et~al.(2020)Aledhari, Razzak, Parizi, and Saeed]{aledhari2020federated}
Mohammed Aledhari, Rehma Razzak, Reza~M Parizi, and Fahad Saeed.
\newblock Federated learning: A survey on enabling technologies, protocols, and applications.
\newblock \emph{IEEE Access}, 8:\penalty0 140699--140725, 2020.

\bibitem[Arivazhagan et~al.(2019)Arivazhagan, Aggarwal, Singh, and Choudhary]{arivazhagan2019federated}
Manoj~Ghuhan Arivazhagan, Vinay Aggarwal, Aaditya~Kumar Singh, and Sunav Choudhary.
\newblock Federated learning with personalization layers, 2019.

\bibitem[Chen et~al.(2019)Chen, Luo, Dong, Li, and He]{chen2019federated}
Fei Chen, Mi Luo, Zhenhua Dong, Zhenguo Li, and Xiuqiang He.
\newblock Federated meta-learning with fast convergence and efficient communication, 2019.

\bibitem[Collins et~al.(2021)Collins, Hassani, Mokhtari, and Shakkottai]{collins2021exploiting}
Liam Collins, Hamed Hassani, Aryan Mokhtari, and Sanjay Shakkottai.
\newblock Exploiting shared representations for personalized federated learning.
\newblock In \emph{International Conference on Machine Learning}, pages 2089--2099. PMLR, 2021.

\bibitem[Deng et~al.(2009)Deng, Dong, Socher, Li, Li, and Fei-Fei]{5206848}
Jia Deng, Wei Dong, Richard Socher, Li-Jia Li, Kai Li, and Li Fei-Fei.
\newblock Imagenet: A large-scale hierarchical image database.
\newblock In \emph{2009 IEEE Conference on Computer Vision and Pattern Recognition}, pages 248--255, 2009.

\bibitem[Duan et~al.(2021)Duan, Liu, Ji, Wu, Liang, Chen, and Tan]{Duan2021FlexibleCF}
Moming Duan, Duo Liu, Xinyuan Ji, Yu Wu, Liang Liang, Xianzhang Chen, and Yujuan Tan.
\newblock Flexible clustered federated learning for client-level data distribution shift.
\newblock \emph{IEEE Transactions on Parallel and Distributed Systems}, 33:\penalty0 2661--2674, 2021.

\bibitem[Fallah et~al.(2020)Fallah, Mokhtari, and Ozdaglar]{fallah2020personalized}
Alireza Fallah, Aryan Mokhtari, and Asuman Ozdaglar.
\newblock Personalized federated learning: A meta-learning approach, 2020.

\bibitem[Frankle and Carbin(2019)]{frankle2019lottery}
Jonathan Frankle and Michael Carbin.
\newblock The lottery ticket hypothesis: Finding sparse, trainable neural networks.
\newblock In \emph{International Conference on Learning Representations}, 2019.

\bibitem[Frankle et~al.(2020)Frankle, Dziugaite, Roy, and Carbin]{frankle2020linear}
Jonathan Frankle, Gintare~Karolina Dziugaite, Daniel~M. Roy, and Michael Carbin.
\newblock Linear mode connectivity and the lottery ticket hypothesis, 2020.

\bibitem[Ghosh et~al.(2020)Ghosh, Chung, Yin, and Ramchandran]{Ghosh2020AnEF}
Avishek Ghosh, Jichan Chung, Dong Yin, and Kannan Ramchandran.
\newblock An efficient framework for clustered federated learning.
\newblock \emph{IEEE Transactions on Information Theory}, 68:\penalty0 8076--8091, 2020.

\bibitem[He et~al.(2015)He, Zhang, Ren, and Sun]{he2015deep}
Kaiming He, Xiangyu Zhang, Shaoqing Ren, and Jian Sun.
\newblock Deep residual learning for image recognition, 2015.

\bibitem[Hendrycks and Dietterich(2019)]{hendrycks2019robustness}
Dan Hendrycks and Thomas Dietterich.
\newblock Benchmarking neural network robustness to common corruptions and perturbations.
\newblock \emph{Proceedings of the International Conference on Learning Representations}, 2019.

\bibitem[Hsu et~al.(2020)Hsu, Qi, and Brown]{hsu2020federated}
Tzu-Ming~Harry Hsu, Hang Qi, and Matthew Brown.
\newblock Federated visual classification with real-world data distribution, 2020.

\bibitem[Jiang et~al.(2023)Jiang, Konečný, Rush, and Kannan]{jiang2023improving}
Yihan Jiang, Jakub Konečný, Keith Rush, and Sreeram Kannan.
\newblock Improving federated learning personalization via model agnostic meta learning, 2023.

\bibitem[Kirkpatrick et~al.(2017)Kirkpatrick, Pascanu, Rabinowitz, Veness, Desjardins, Rusu, Milan, Quan, Ramalho, Grabska-Barwinska, Hassabis, Clopath, Kumaran, and Hadsell]{Kirkpatrick_2017}
James Kirkpatrick, Razvan Pascanu, Neil Rabinowitz, Joel Veness, Guillaume Desjardins, Andrei~A. Rusu, Kieran Milan, John Quan, Tiago Ramalho, Agnieszka Grabska-Barwinska, Demis Hassabis, Claudia Clopath, Dharshan Kumaran, and Raia Hadsell.
\newblock Overcoming catastrophic forgetting in neural networks.
\newblock \emph{Proceedings of the National Academy of Sciences}, 114\penalty0 (13):\penalty0 3521--3526, 2017.

\bibitem[Krizhevsky(2009)]{Krizhevsky2009LearningML}
Alex Krizhevsky.
\newblock Learning multiple layers of features from tiny images.
\newblock 2009.

\bibitem[Lee et~al.(2023)Lee, Chen, Tajwar, Kumar, Yao, Liang, and Finn]{lee2023surgical}
Yoonho Lee, Annie~S. Chen, Fahim Tajwar, Ananya Kumar, Huaxiu Yao, Percy Liang, and Chelsea Finn.
\newblock Surgical fine-tuning improves adaptation to distribution shifts, 2023.

\bibitem[Li et~al.(2020{\natexlab{a}})Li, Sun, Wang, Duan, Li, Chen, and Li]{li2020lotteryfl}
Ang Li, Jingwei Sun, Binghui Wang, Lin Duan, Sicheng Li, Yiran Chen, and Hai Li.
\newblock Lotteryfl: Personalized and communication-efficient federated learning with lottery ticket hypothesis on non-iid datasets, 2020{\natexlab{a}}.

\bibitem[Li et~al.(2019)Li, Wen, Wu, and He]{Li2019ASO}
Q. Li, Zeyi Wen, Zhaomin Wu, and Bingsheng He.
\newblock A survey on federated learning systems: Vision, hype and reality for data privacy and protection.
\newblock \emph{IEEE Transactions on Knowledge and Data Engineering}, 35:\penalty0 3347--3366, 2019.

\bibitem[Li et~al.(2020{\natexlab{b}})Li, Sahu, Talwalkar, and Smith]{Li_2020}
Tian Li, Anit~Kumar Sahu, Ameet Talwalkar, and Virginia Smith.
\newblock Federated learning: Challenges, methods, and future directions.
\newblock \emph{{IEEE} Signal Processing Magazine}, 37\penalty0 (3):\penalty0 50--60, 2020{\natexlab{b}}.

\bibitem[Li et~al.(2020{\natexlab{c}})Li, Sahu, Zaheer, Sanjabi, Talwalkar, and Smith]{li2020federated}
Tian Li, Anit~Kumar Sahu, Manzil Zaheer, Maziar Sanjabi, Ameet Talwalkar, and Virginia Smith.
\newblock Federated optimization in heterogeneous networks, 2020{\natexlab{c}}.

\bibitem[Li et~al.(2021{\natexlab{a}})Li, Hu, Beirami, and Smith]{li2021ditto}
T. Li, S. Hu, A. Beirami, and V. Smith.
\newblock Ditto: Fair and robust federated learning through personalization.
\newblock In \emph{International Conference on Machine Learning}, pages 6357--6368. PMLR, 2021{\natexlab{a}}.

\bibitem[Li et~al.(2021{\natexlab{b}})Li, JIANG, Zhang, Kamp, and Dou]{lifedbn}
Xiaoxiao Li, Meirui JIANG, Xiaofei Zhang, Michael Kamp, and Qi Dou.
\newblock Fedbn: Federated learning on non-iid features via local batch normalization.
\newblock In \emph{International Conference on Learning Representations}, 2021{\natexlab{b}}.

\bibitem[Liang et~al.(2020)Liang, Liu, Ziyin, Allen, Auerbach, Brent, Salakhutdinov, and Morency]{liang2020think}
Paul~Pu Liang, Terrance Liu, Liu Ziyin, Nicholas~B Allen, Randy~P Auerbach, David Brent, Ruslan Salakhutdinov, and Louis-Philippe Morency.
\newblock Think locally, act globally: Federated learning with local and global representations.
\newblock \emph{arXiv preprint arXiv:2001.01523}, 2020.

\bibitem[Liu et~al.(2022)Liu, Hu, Wu, and Smith]{liu2022privacy}
Ziyu Liu, Shengyuan Hu, Zhiwei~Steven Wu, and Virginia Smith.
\newblock On privacy and personalization in cross-silo federated learning, 2022.

\bibitem[Mansour et~al.(2020)Mansour, Mohri, Ro, and Suresh]{mansour2020approaches}
Yishay Mansour, Mehryar Mohri, Jae Ro, and Ananda~Theertha Suresh.
\newblock Three approaches for personalization with applications to federated learning, 2020.

\bibitem[McCloskey and Cohen(1989)]{McCloskey1989CatastrophicII}
Michael McCloskey and Neal~J. Cohen.
\newblock Catastrophic interference in connectionist networks: The sequential learning problem.
\newblock \emph{Psychology of Learning and Motivation}, 24:\penalty0 109--165, 1989.

\bibitem[McMahan et~al.(2017)McMahan, Moore, Ramage, Hampson, and y~Arcas]{mcmahan2017}
Brendan McMahan, Eider Moore, Daniel Ramage, Seth Hampson, and Blaise~Aguera y Arcas.
\newblock Communication-efficient learning of deep networks from decentralized data.
\newblock In \emph{Proc. of Int’l Conf. Artificial Intelligence and Statistics (AISTATS)}, 2017.

\bibitem[Mugunthan et~al.(2022)Mugunthan, Lin, Gokul, Lau, Kagal, and Pieper]{10.1007/978-3-031-19775-8_5}
Vaikkunth Mugunthan, Eric Lin, Vignesh Gokul, Christian Lau, Lalana Kagal, and Steve Pieper.
\newblock Fedltn: Federated learning for sparse and personalized lottery ticket networks.
\newblock In \emph{Computer Vision -- ECCV 2022}, pages 69--85, Cham, 2022. Springer Nature Switzerland.

\bibitem[Oh et~al.(2022)Oh, Kim, and Yun]{oh2022fedbabu}
Jaehoon Oh, SangMook Kim, and Se-Young Yun.
\newblock Fed{BABU}: Toward enhanced representation for federated image classification.
\newblock In \emph{International Conference on Learning Representations}, 2022.

\bibitem[Pillutla et~al.(2022)Pillutla, Malik, Mohamed, Rabbat, Sanjabi, and Xiao]{pillutla2022federated}
K. Pillutla, K. Malik, A. Mohamed, M. Rabbat, M. Sanjabi, and L. Xiao.
\newblock Federated learning with partial model personalization.
\newblock In \emph{International Conference on Machine Learning}, 2022.

\bibitem[Renda et~al.(2020)Renda, Frankle, and Carbin]{renda2020comparing}
Alex Renda, Jonathan Frankle, and Michael Carbin.
\newblock Comparing rewinding and fine-tuning in neural network pruning, 2020.

\bibitem[Shamsian et~al.(2021)Shamsian, Navon, Fetaya, and Chechik]{shamsian2021personalized}
Aviv Shamsian, Aviv Navon, Ethan Fetaya, and Gal Chechik.
\newblock Personalized federated learning using hypernetworks.
\newblock In \emph{International Conference on Machine Learning}, pages 9489--9502. PMLR, 2021.

\bibitem[Sheller et~al.(2020)Sheller, Edwards, Reina, Martin, and Bakas]{sheller2020federated}
M.~J. Sheller, B. Edwards, G.~A. Reina, J. Martin, and S. Bakas.
\newblock Federated learning in medicine: facilitating multi-institutional collaborations without sharing patient data.
\newblock \emph{Scientific Reports}, 10\penalty0 (12598), 2020.

\bibitem[Smith et~al.(2017)Smith, Chiang, Sanjabi, and Talwalkar]{Smith2017FederatedML}
Virginia Smith, Chao-Kai Chiang, Maziar Sanjabi, and Ameet Talwalkar.
\newblock Federated multi-task learning.
\newblock \emph{ArXiv}, abs/1705.10467, 2017.

\bibitem[Sun et~al.(2021)Sun, Huo, Yang, and Bai]{sun2021partialfed}
Benyuan Sun, Hongxing Huo, Yi Yang, and Bo Bai.
\newblock Partialfed: Cross-domain personalized federated learning via partial initialization.
\newblock \emph{Advances in Neural Information Processing Systems}, 34:\penalty0 23309--23320, 2021.

\bibitem[Tan et~al.(2022)Tan, Yu, Cui, and Yang]{tan2022towards}
Alysa~Ziying Tan, Han Yu, Lizhen Cui, and Qiang Yang.
\newblock Towards personalized federated learning.
\newblock \emph{IEEE Transactions on Neural Networks and Learning Systems}, 2022.

\bibitem[Venkateswara et~al.(2017)Venkateswara, Eus{\'e}bio, Chakraborty, and Panchanathan]{Venkateswara2017DeepHN}
Hemanth Venkateswara, Jos{\'e} Eus{\'e}bio, Shayok Chakraborty, and Sethuraman Panchanathan.
\newblock Deep hashing network for unsupervised domain adaptation.
\newblock \emph{2017 IEEE Conference on Computer Vision and Pattern Recognition (CVPR)}, 2017.

\bibitem[Vinyals et~al.(2017)Vinyals, Blundell, Lillicrap, Kavukcuoglu, and Wierstra]{vinyals2017matching}
Oriol Vinyals, Charles Blundell, Timothy Lillicrap, Koray Kavukcuoglu, and Daan Wierstra.
\newblock Matching networks for one shot learning.
\newblock In \emph{Neural Information Processing Systems}, 2017.

\bibitem[Wang et~al.(2020)Wang, Xu, Wang, and Zhu]{wang2020addressing}
Lixu Wang, Shichao Xu, Xiao Wang, and Qi Zhu.
\newblock Addressing class imbalance in federated learning, 2020.

\bibitem[Xu et~al.(2023)Xu, Tong, and Huang]{xu2023personalized}
Jian Xu, Xinyi Tong, and Shao-Lun Huang.
\newblock Personalized federated learning with feature alignment and classifier collaboration.
\newblock In \emph{The Eleventh International Conference on Learning Representations}, 2023.

\bibitem[Yosinski et~al.(2014)Yosinski, Clune, Bengio, and Lipson]{Yosinski2014HowTA}
Jason Yosinski, Jeff Clune, Yoshua Bengio, and Hod Lipson.
\newblock How transferable are features in deep neural networks?
\newblock In \emph{Neural Information Processing Systems}, 2014.

\bibitem[Yu et~al.(2022)Yu, Bagdasaryan, and Shmatikov]{yu2022salvaging}
Tao Yu, Eugene Bagdasaryan, and Vitaly Shmatikov.
\newblock Salvaging federated learning by local adaptation, 2022.

\bibitem[Zhao et~al.(2018)Zhao, Li, Lai, Suda, Civin, and Chandra]{zhao2018iid}
Yue Zhao, Meng Li, Liangzhen Lai, Naveen Suda, Damon Civin, and Vikas Chandra.
\newblock Federated learning with non-iid data.
\newblock 2018.

\end{thebibliography}
}

\end{document}


\maketitle


\section{Experimental Setup}\label{app:ep-details}

In this section, we describe the details of our experimental setup used for the results in the main text. We include information about our ablation study as well as hyperparameters used for training both \textsc{\algname{}} and the compared baselines. 

\subsection{Implementation Details}
\label{sec:implementation_details}

\paragraph{Models \& Datasets.} We conduct experiments on 4 datasets: CIFAR-10  \citep{Krizhevsky2009LearningML}, CIFAR10-C \citep{hendrycks2019robustness}, Mini-ImageNet \citep{vinyals2017matching}, and OfficeHome \citep{Venkateswara2017DeepHN}. CIFAR10 contains 60,000 image samples across 10 labeled image categories, from which 50,000 are selected for training and the remaining 10,000 for testing; CIFAR10-C contains the same set of training samples as CIFAR10, with different types of image corruptions (severity $=5$) for each client. Mini-ImageNet contains 50,000 training samples and 10,000 testing samples, across 100 image categories. Finally, the OfficeHome dataset contains data across 4 domain shifts and 65 image categories with 2162, 3909, 3969, 3892 training samples for each of the 4 domains, respectively. The corresponding test set sizes are 265, 456, 470, and 465 testing samples.

\paragraph{Data Partioning} For all experiments, training samples were distributed to each client in a non-IID fashion. For experiments involving either CIFAR10 or CIFAR10-C, training samples for each client were additionally truncated to a fixed size $N_k$; however the test set size for each client remained fixed at 200. We follow the learning setups presented in \citep{li2020lotteryfl, xu2023personalized} and limit the number of data samples per client to ensure the necessity of federated learning to achieve optimal performance as opposed to pure local training. We define the shard $s$ as the number of classes assigned to a client, which is set to $s=2$ in all CIFAR10 \& CIFAR10-C experiments and $s=10$ for the Mini-ImageNet experiment. For the OfficeHome experiment, there are 4 clients for each domain shift, such that each client is allocated to the full dataset of its domain as in \cite{sun2021partialfed}.

\paragraph{Types of Distributional Shift.} For the CIFAR10 and Mini-ImageNet experiments, we use a label-based partition; therefore there is label shift among the clients' data distributions. For the CIFAR10-C experiments, while using a labeled-based partition, we also apply a random type of image corruption to each client, which introduces feature shift. Finally for the OfficeHome experiments, there is primarily feature shift since the images from each client are from different domains.

\paragraph{Compared Algorithms.} In addition to FedAvg \citep{mcmahan2017}, we compare our method to several \textit{full model personalization} methods: local-only training, FedAvg with local fine-tuning (FedAvg + FT), and Ditto \cite{li2021ditto} which personalizes clients via a multi-task learning objective. We also compare against \textit{partial model personalization} methods:  FedBABU \cite{oh2022fedbabu} which only updates and aggregates the feature extractor; FedRep \cite{collins2021exploiting} and FedPer \cite{arivazhagan2019federated} which aggregate client feature extractors and personalize classifier heads; LG-FedAvg \citep{liang2020think} which personalizes the clients' feature extractors; FedPAC \cite{xu2023personalized} which performs local-global feature alignment via classifier collaboration.

\paragraph{Hyperparameters. } For all methods, we use 3 local training epochs using stochastic gradient descent (SGD) optimizer momentum. We tuned the learning rate over $\{0.1, 0.01, 0.001\}$ for all of the compared baselines. For the parameter decoupling methods \citep{oh2022fedbabu, collins2021exploiting, arivazhagan2019federated, liang2020think}, this entails setting the learning rates for the respective feature extractors and classifier heads that are locally updated. As a result, we used a learning rate of $0.01$ for the CIFAR-10 and CIFAR10-C experiments, and a learning rate of $0.001$ for the OfficeHome and Mini-ImageNet experiments. For \textsc{\algname{}}, we follow the recommendation of choosing different learning rates for global and personal parameters given in the partial model personalization framework \citep{pillutla2022federated}. Specifically, we use a learning rate of $0.1$ for the personal parameters and $0.001$ for the global parameters for the CIFAR-10 experiments. Our reported experiments (excluding Table \ref{tab:tablevarper}) use a personalization rate of $p=0.05$. Extended results for a grid search across a selection of values for $p$ and $\alpha$ are described in Appendix \ref{sec:effectpa}. We set the personalization hyperparameter $\lambda$ of Ditto to $0.75$ after tuning over $\{0.25, 0.50, 0.75\}$, which follows the setting used by FedBABU \cite{oh2022fedbabu}.

\section{Additional Experimental Results}\label{app:ep-more-results}

In this section we provide further experimental results and clarify additional details of the ablation study described in the main text.

\subsection{Ablation Study Details} One of the components of our ablation study of the key design choices involved in \textsc{\algname{}} involves choosing 4 different layers (denoted as Layer A/B/C/D in Section 5.2 of the main text) for personalization. We selected Layers A, B, C, and D from ResNet18 given by their PyTorch \citep{paszke2019pytorch} layer-names in the library-provided ResNet18 implementation: Layer A refers to 'conv1.weight'; Layer B refers to 'layer1.0.conv1.weight'; Layer C refers to 'layer2.0.conv1.weight'; Layer D refers to 'layer4.1.conv1.weight'.

\subsection{Effect of $p$ and $\alpha$ on personalization}
For the following described experiments in this subsection, we use the same data partition split as in the CIFAR-10 experiments presented in the main text; 100 training samples and 200 testing samples are allocated for each client with 10 clients undergoing full participation FL, and each client is allocated 2 image category classes.

\label{sec:effectpa}

\paragraph{Varying the personalization rate $p$.} We present additional experimental results for varying the personalization rate $p$ alongside the personalization limit $\alpha$ for CIFAR-10 in Table \ref{tab:tablevarper}.  We observe the best performance occurs when $p=0.05$ for a limit of $\alpha=0.30$. We also note that when $p > \alpha$, the personalization limit is reached after the first round. In general, we suggest using a smaller $p$ with a larger $\alpha$ when attempting to personalize many client parameters, rather than personalizing with a very high rate ($p > 0.20$).

\begin{table}
  \centering
  \scriptsize
  \setlength{\tabcolsep}{1.0mm}{
  \begin{tabular}{@{}lcccc@{}}
    \toprule
     & $\alpha=0.05$ & $\alpha=0.30$ & $\alpha=0.50$ & $\alpha=0.80$ \\
    \midrule
     $p=0.01$ & 81.35 & 81.35 & 81.85 & 80.65 \\ 
     $p=0.05$ & 80.10 & 82.25 & 82.20 & 81.40 \\
     $p=0.20$ & 80.55 & 81.70 & 81.65 & 82.15 \\
     $p=0.50$ & 82.05 & 82.05 & 81.60 & 81.20 \\
    \bottomrule
  \end{tabular}
  }
  \caption{Performance (\% mean client test accuracy) of \textsc{\algname{}} on CIFAR-10 when varying both the personalization limit $\alpha$ and the personalization rate $p$.}
  \label{tab:tablevarper}
   \vspace{-4mm}
\end{table}

\paragraph{Varying the personalization limit $\alpha$.} We showcase the performance of \textsc{\algname{}} on CIFAR-10 under a wide range of values for $\alpha \in [0, 1]$ in Table \ref{tab:tablevarperextend} for a fixed personalization rate $p=0.05$. From Section 4.4 of the main text, we have that \textsc{\algname{}} intuitively performs an interpolation of the two extremes of personalization: pure federated averaging ($\alpha=0.0$) and eventual pure local training $\alpha=1.0$. We generally recommend using middle-ground values of $\alpha \in [0.3, 0.5]$.

\begin{table}
  \centering
  \scriptsize
  \setlength{\tabcolsep}{1.0mm}{
  \begin{tabular}{@{}lcc@{}}
    \toprule
    $\alpha$ & CIFAR10 & CIFAR10-C \\
    \midrule
     0.05 & 80.10 & 68.30 \\
     0.10 & 81.25 & 69.05  \\ 
     0.20 & 81.15 & 70.65  \\
     0.30 & \textbf{82.25} & 70.60  \\
     0.40 & 81.10 & 70.55  \\
     0.50 & 82.20 & \textbf{72.05}  \\
     0.60 & 81.60 & 70.90  \\
     0.80 & 81.40 & 71.40 \\
    \bottomrule
  \end{tabular}
  }
  \caption{Performance (\% mean client test accuracy) of \textsc{\algname{}} on both CIFAR-10 and CIFAR-10C when varying the personalization limit $\alpha$.}
  \label{tab:tablevarperextend}
   \vspace{-4mm}
\end{table}



\subsection{Robustness to sample size.} We present the mean client performance of \textsc{\algname{}} and the compared baselines for client training data sizes $N_k \in \{20, 40, 100, 200\}$ in Table \ref{tab:vardatasize}, where $k$ represents the $k$-th client $c_k$. To ensure a fair comparison of performance across each setting of $N_k$, we designated the same test sets for each client, which had 200 testing samples each. Notably, we observe that \textsc{\algname{}} maintains superior performance across each data size setting, particularly for low client training dataset sizes $N_k \leq 40$, where algorithms such as FedBABU, Ditto, and FedPer suffer a significant performance decrease. 

\begin{table}
  \centering
  \scriptsize
    \setlength{\tabcolsep}{1.0mm}{
    \begin{tabular}{@{}lcccc@{}}
    \toprule
    Method & $N_k=20$ & $N_k=40$ & $N_k=100$ & $N_k=200$\\
    \midrule 
    FedAvg & 23.55 & 24.70 & 27.70 & 26.90\\
    FedAvg + FT & 37.65 & 72.35 & 75.30 & 83.50 \\
    FedPAC & 68.30 & 67.00 & 77.20 & 81.65 \\
    FedRep  & 70.10 & 73.00 & 67.60 & 78.15  \\
    FedPer & 55.70 & 75.15 & 75.40 & 83.15 \\
    FedBABU & 33.10 & 66.10 & 75.45 & 81.80\\
    Ditto & 37.25 & 71.55 & 72.75 & 81.65\\
    LG-FedAvg & 71.45 & 74.70 & 77.65 & 83.90\\
    \algname{} & \textbf{72.25} & \textbf{78.20} & \textbf{82.25} & \textbf{84.85}   \\
    \bottomrule
  \end{tabular}
  }
  \caption{Comparison of performance (\% mean client test accuracy) on CIFAR-10 when varying the training data size $N_k$
 per client $c_k$. Each client was assigned with 2 classes.}
  \label{tab:vardatasize}
\end{table}

\newpage
\section{Convergence Analysis}

In this section, we analyze the convergence of \textsc{\algname{}}. Specifically, we first prove in Theorem \ref{converge:masks} that the mask for each client is guaranteed to converge in finite communication rounds. We then prove in Theorem \ref{converge:params} that once the masks for all clients converge, the model parameters have the same convergence guarantee as block stochastic gradient descent. 

\begin{theorem}[Convergence of masks] \label{converge:masks}
    Under full client participation, the masks for all clients converge in finite communication rounds. 
\end{theorem}

\begin{proof}
    For each client $c_k$, the set of personalized parameter indices are monotonically increasing, i.e., $\text{Idx}(v_k^0) \subset \text{Idx}(v_k^1) \subset \cdots \subset \text{Idx}(v_k^T)$. Meanwhile, the cardinality of this set is upper bounded by $\alpha d$, where $\alpha$ is the personalization limit and $d$ is the dimensionality of the parameter. Therefore, the masks for all clients converge in finite communication rounds. 
\end{proof}

\begin{theorem}[Convergence of parameters] \label{converge:params}
    Under full client participation, when the number of local steps $\tau = 1$ and participation, after the masks for all clients converge, \textsc{\algname{}} has the identical convergence guarantee as centralized block stochastic gradient descent (block SGD). 
\end{theorem}

\begin{proof}
    We let $\theta_k[i]$ denote the $i$-th parameter on client $c_k$ and $m_k[i]$ denote its corresponding mask, where $m_k[i] = 0$ means the parameter is globally shared and $m_k[i] = 1$ means the parameter is personalized (on client $c_k$). If a parameter is global, i.e., shared across several clients, we use $\theta[i]$ to denote such parameter. After the masks for all clients converge, \textsc{\algname{}} becomes an optimization problem over the union of the global parameters and each client's personalized parameters: 
    \begin{itemize}
        \item Global parameters: $U = \{\theta[i] : \forall i, \exists k, \text{ s.t. } m_k[i] = 0\}$
        \item Personalized parameters: $V = \{\theta_k[i]: \forall k, i, \text{ s.t. } m_k[i] = 1\}$
    \end{itemize}
    and the optimization objective is
    \begin{align*}
        F(U, V) = \frac{1}{N} \sum_{k=1}^N f_k(u_k, v_k)
    \end{align*}
    where $u_k \subset U$ and $\cup_{k=1}^N v_k = V$. 
    
    For example, consider a system with 3 clients and corresponding masks 
    \begin{align*}
        m_1 = [1, 1, 0, 0], m_2 = [1, 0, 1, 0], m_3 = [1, 0, 0, 1]
    \end{align*}
    then, we have
    \begin{align*}
        u_1 &= [\theta_1[3], \theta_1[4]] = [\theta[3], \theta[4]], \quad v_1 = [\theta_1[1], \theta_1[2]], \\
        u_2 &= [\theta_2[2], \theta_1[4]] = [\theta[2], \theta[4]], \quad v_2 = [\theta_2[1], \theta_2[3]], \\
        u_3 &= [\theta_3[2], \theta_3[3]] = [\theta[2], \theta[3]], \quad v_3 = [\theta_3[1], \theta_3[4]], 
    \end{align*}
    and 
    \begin{align*}
        U &= [ \theta[2], \theta[3], \theta[4] ], \\
        V &= [\theta_1[1], \theta_1[2], \theta_2[1], \theta_2[3], \theta_3[1], \theta_3[4]]
    \end{align*}
    For clarification, we use the original client and parameter indices $k$ and $i$ as the indices of $U$ and $V$. 
    
    After the masks converge, in each communication round, each client conducts LocalAlt (see Algorithm 2), which is SGD on personalized parameters $v$ followed by SGD on global parameters $u$. After LocalAlt, the global parameters are aggregated on the server. Next, we show that each communication round is equivalent to central block SGD on $U$ and $V$ alternatively. 
    \begin{itemize}
        \item Update personalized parameters: Each client independently optimize its personalized parameter. This is equivalent to centralized SGD w.r.t. $V$, since
        \begin{align*}
            \frac{\partial F}{\partial V_k[i]} = \frac{\partial F}{\partial \theta_k[i]}
        \end{align*}

        \item Update global parameters: Eech client first optimize its local copy global parameter. Then, each client's updated local copy will be uploaded to the server for aggregation, i.e., for the $i$-th parameter, 
        \begin{align*}
            u_k^+[i] &\leftarrow u_k[i] - \gamma_u \frac{\partial f_k(u_k, v_k)}{\partial u_k[i]}, \forall k \tag{local update} \\
            u_k[i] &\leftarrow \frac{1}{\sum_{k=1}^N (1 - m_k[i])}\sum_{k=1}^N (1 - m_k[i]) u_k^+[i]  \tag{aggregation}
        \end{align*}
        This is equivalent to 
        \begin{align*}
            u_k[i] &\leftarrow u_k[i] - \left( \gamma_u \frac{N}{\sum_{k=1}^N (1 - m_k[i])} \right) \cdot \frac{\partial F}{\partial U_k[i]}
        \end{align*}
        which is centralized SGD w.r.t. $U$. 
    \end{itemize}
    Therefore, the optimization process is numerically equivalent to block SGD with $U, V$ alternatively. 
\end{proof}

\newpage
{
    \small
    \bibliographystyle{ieeenat_fullname}
    \bibliography{main}
}